\documentclass[10pt,twocolumn,letterpaper]{article}

\usepackage[pagenumbers]{cvpr} 

\usepackage{times}
\usepackage{epsfig}
\usepackage{graphicx}
\usepackage{amsmath}
\usepackage{amssymb}
\usepackage{booktabs}
\usepackage{arydshln}
\usepackage{wrapfig}
\usepackage{hhline}
\usepackage{array}
\usepackage{bm}
\usepackage{multirow}
\usepackage[table,xcdraw]{xcolor}
\usepackage{diagbox}
\usepackage{rotfloat}
\usepackage{CJKutf8}
\usepackage{verbatim}
\usepackage{pifont}
\usepackage{color}
\usepackage{overpic}
\usepackage{makecell}
\usepackage[T1]{fontenc}
\usepackage{enumitem}
\usepackage{colortbl}
\usepackage{overpic}

\definecolor{darkpastelgreen}{rgb}{0.01, 0.75, 0.24}
\definecolor{darkpink}{rgb}{0.91, 0.33, 0.5}
\definecolor{mygray}{gray}{.92}

\definecolor{linkcolor}{RGB}{255,0,0}
\definecolor{urlcolor}{RGB}{255,105,180}
\definecolor{citecolor}{RGB}{0, 80, 200}
\definecolor{citecolor1}{RGB}{0,153,255}
\usepackage{hyperref}
\hypersetup{colorlinks=true,linkcolor=linkcolor,urlcolor=urlcolor,citecolor=citecolor1}

\newcolumntype{x}{>{\columncolor[HTML]{EFEFEF}[16pt]}r}
\newcolumntype{y}{>{\columncolor[HTML]{EFEFEF}[0pt]}l}

\usepackage{algorithm}
\usepackage{algpseudocode}
\algnewcommand\algorithmicinput{\textbf{Input:}}
\newlength{\maxwidth}
\newcommand{\algalign}[2]
{\makebox[\maxwidth][l]{$#1{}$}${}#2$}

\usepackage{multicol}
\usepackage{setspace}
\usepackage{caption}

\usepackage[capitalize]{cleveref}
\crefname{section}{Sec.}{Secs.}
\Crefname{section}{Section}{Sections}
\Crefname{table}{Table}{Tables}
\crefname{table}{Tab.}{Tabs.}

\def\cvprPaperID{544} 
\def\confName{CVPR}
\def\confYear{2022}

\begin{document}

\title{Self-Sustaining Representation Expansion for \\
Non-Exemplar Class-Incremental Learning}
\author{Kai~Zhu\textsuperscript{\rm 1}\qquad~Wei~Zhai\textsuperscript{\rm 1}\qquad~Yang~Cao\textsuperscript{\rm 1,}\textsuperscript{\rm 3,}\footnotemark[2]\qquad~Jiebo~Luo\textsuperscript{\rm 2}\qquad~Zheng-Jun~Zha\textsuperscript{\rm 1}\\
		{\textsuperscript{\rm 1} University of Science and Technology of China}  \qquad
		{\textsuperscript{\rm 2} University of Rochester} \\
		{\textsuperscript{\rm 3} Institute of Artificial Intelligence, Hefei Comprehensive National Science Center} \\
		\small{\texttt{\{zkzy@mail., wzhai056@mail., forrest@\}ustc.edu.cn}} \quad
		\small{\texttt{jluo@cs.rochester.edu}} \quad
		\small{\texttt{zhazj@ustc.edu.cn}}
	}
	
\maketitle
\renewcommand{\thefootnote}{\fnsymbol{footnote}}
\footnotetext[2]{Corresponding Author}

\begin{abstract}
  Non-exemplar class-incremental learning is to recognize both the old and new classes when old class samples 
  cannot be saved. It is a challenging task since representation optimization and feature retention  
  can only be achieved under supervision from new classes. To address this problem, we propose a novel self-sustaining
  representation expansion scheme. Our scheme consists of a structure reorganization strategy that fuses main-branch expansion and side-branch updating to maintain the old features, and a main-branch distillation scheme to transfer the invariant knowledge. Furthermore, a  prototype selection  mechanism is proposed to enhance the discrimination between the old and new classes by selectively incorporating new samples into the distillation process. Extensive experiments on three benchmarks demonstrate significant incremental performance, outperforming the state-of-the-art methods by a margin of $3\% $, $3\% $ and $6\% $, respectively.
\end{abstract}

\section{Introduction}
\label{sec:intro}
Since deep neural networks have made great advances in fully supervised conditions, research attention is increasingly turning to other aspects of learning. An important aspect is the ability to continuously 
  learn new tasks as the input stream is updated, which is often the case in real applications. In recent years, class-incremental learning (CIL) \cite{rebuffi2017icarl, hou2019learning}, a difficult type in continual learning, has attracted much attention, which aims to recognize new classes without forgetting the old ones that have been learned.

\begin{figure}[t]
  \centering
  \includegraphics[width=1\linewidth]{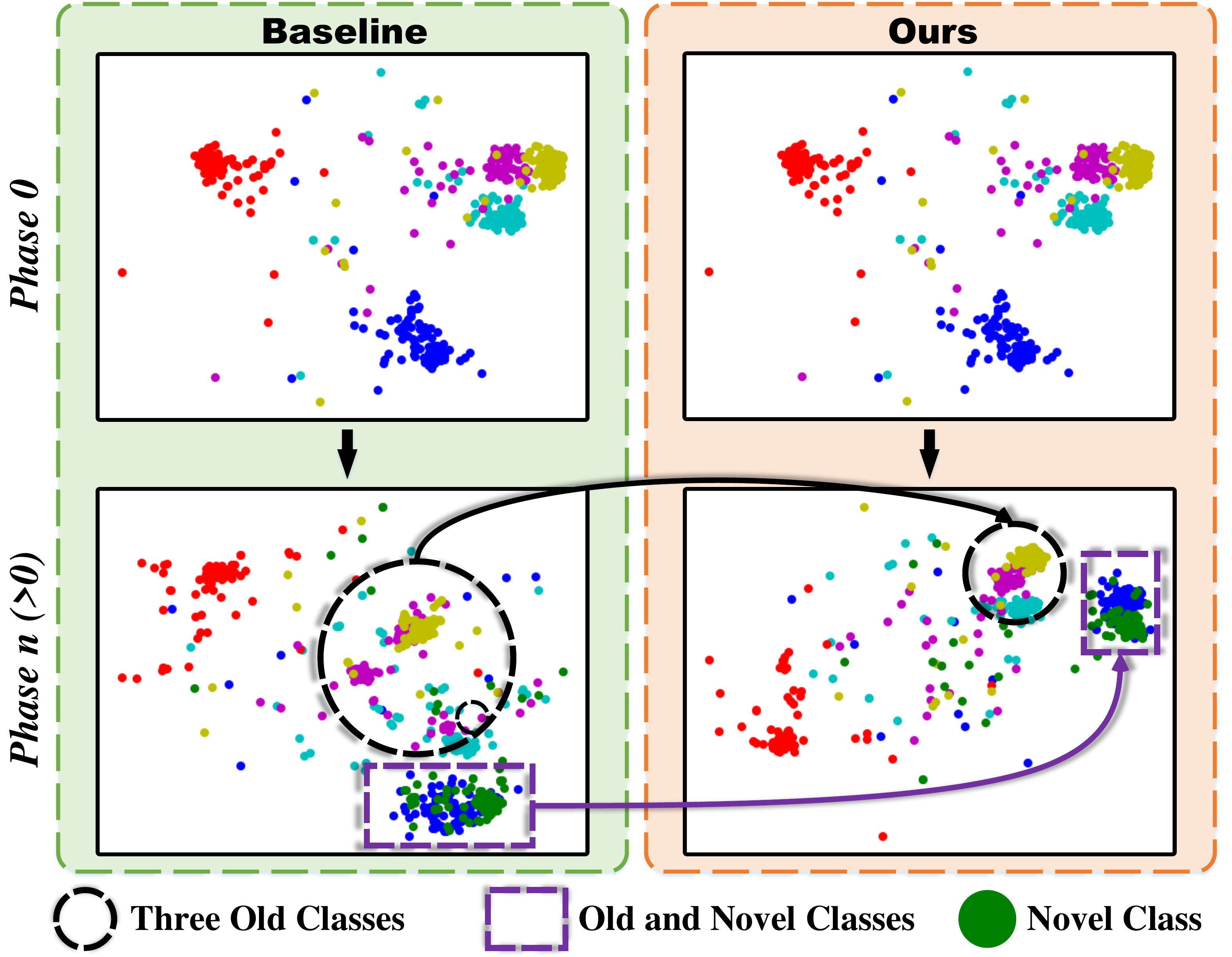}
     \caption{The t-SNE visualization. Compared to the baseline in Section \ref{sec:baseline}, (1) the representations of the old classes in our method are better maintained (circular area), (2) and the novel class is more discriminating from the old classes (rectangular area).}
  \label{fig:first}
\end{figure}

\begin{figure*}[ht]
  \centering
     \includegraphics[width=0.93\linewidth]{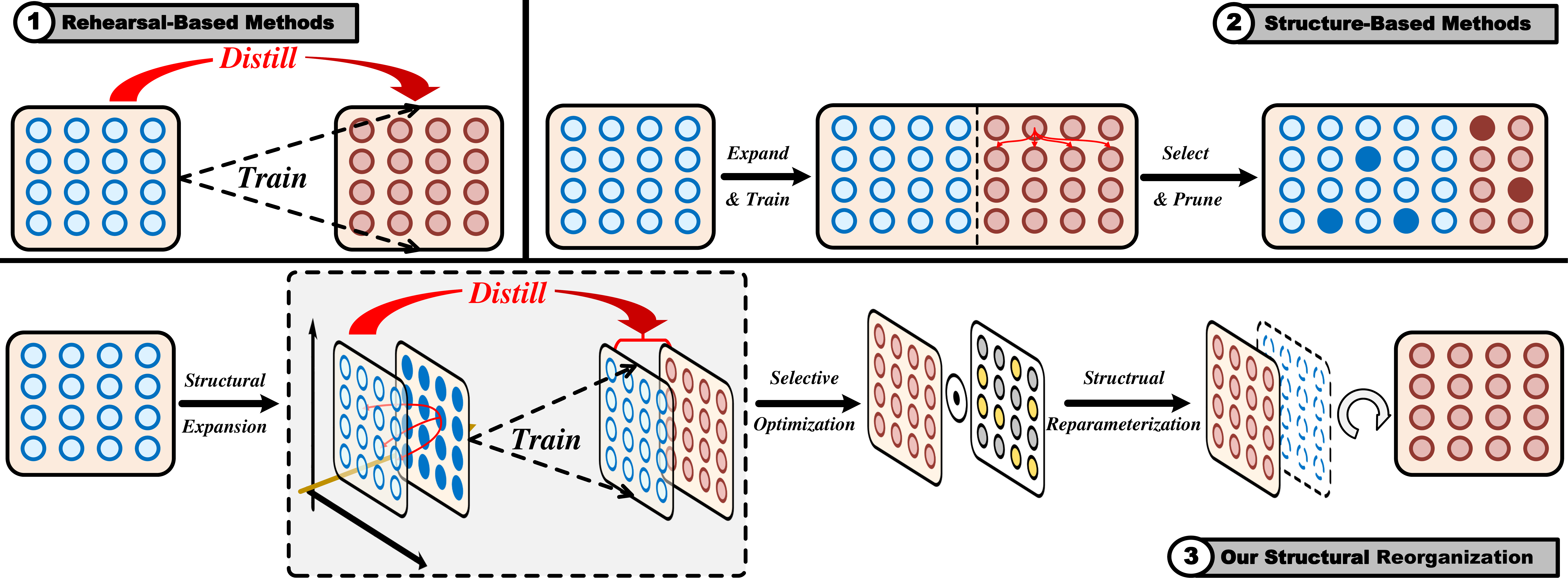}
  \caption{Motivation of our method. In NECIL, the rehearsal-based and structure-based methods suffer from the unreliability of distillation in the absence of exemplars and continuously expanding structure, respectively. DSR is proposed to drive network to expand from a structurally recoverable direction, thus maintaining the discrimination during the new optimization process. On this foundation, we utilize MBD to exploit the ability of distillation-based methods to balance old and new class knowledge.}
  \label{fig:motivation}
\end{figure*}

  In this case, re-training the old and new class samples jointly in each phase is time-consuming and laborious, not to mention that the old class samples may not be fully available. A simple alternative is to fine-tune the network using the new class, however, it will cause the catastrophic forgetting problem \cite{french1999catastrophic}. That is, during the optimization process, the entire representation and the classifier become biased toward the new class, resulting in a sharp drop in the performance for the old class. To deal with it, recent CIL methods maintain the past knowledge by preserving some representative samples (\textit{i.e.}, exemplars \cite{rebuffi2017icarl}) and introducing various distillation losses \cite{douillard2020podnet}, and correct the bias caused by number imbalance by calibrating the classifier \cite{hou2019learning}. 

  However, most of the existing methods \cite{liu2021adaptive, simon2021learning} assume that a certain number (\textit{e.g.}, 2000) of exemplars can be stored in memory, which is usually difficult to satisfy in practice due to user privacy or device limitations. This fact poses great difficulties to incremental learning, because the optimization of the representation and the correction of the classifier will degenerate directly from the imbalance between the old and new classes. To this end, this paper focuses on this ability of incrementally learning new classes where old class samples cannot be preserved, which is called non-exemplar class-incremental learning (NECIL \cite{zhu2021prototype}).

  A natural idea for this problem is to directly transfer the existing CIL framework (\textit{i.e.}, rehearsal-based and structure-based methods in Section \ref{related:incremental}) to NECIL, but the experimental results show that this way leads to performance degradation and parameter explosion. On one hand, in rehearsal-based methods, due to the lack of old class samples, the distillation that the new class samples participate in is the only one that can help maintain the representation of old classes. However, for new samples, it is impractical to provide the same complete old class distribution as the exemplars, so it is difficult to effectively promote the knowledge transfer in the distillation process. Consequently, representative features learned in the old phase are lost phase by phase with the decrease of relevance to the new class.
  
  On the other hand, the idea of structure-based methods is to leave the old model for inference and expand a new model for training at each new phase \cite{rusu2016progressive, yan2021dynamically}. Although this strategy maintains the performance of the old class completely, demonstrating strong performance \cite{yan2021dynamically}, the network parameters that increase linearly with phase (\textit{i.e.}, 5, 10 and 20 in this paper) during training are discouraging. Besides, although a large amount of data can be used to learn the discriminative features among new classes, it is easy to confuse with similar ones from the old distribution. The augmentation of prototypes \cite{zhu2021prototype} can only improve the selection of the optimal boundary for the classifier, but cannot essentially improve the discrimination of the old and new classes in the feature representation. As shown in Fig. \ref{fig:first}, the representations of old classes obtained by the standard CIL method are more confused compared to the initial phase, because they may gradually overlap with similar classes due to the lack of effective supervision. At the same time, the new class may directly overlap with the old cluster, resulting in serious confusion to the subsequent optimization process.

  To address this problem, we propose a self-sustaining representation expansion scheme to learn a structure-cyclic representation, promoting the optimization from the expanded direction while integrating the overall structure at the end of each phase. As shown in Fig. \ref{fig:motivation}, the preservation of the old classes is reflected in both the structure and feature aspects. First, we adopt a dynamic structure reorganization (DSR) strategy, which leaves structured space for the learning of new class while stably preserving the old class space through maintaining heritage at the main-branch and fusing update at the side-branch. Second, on the basis of the expandable structure, we employ a main-branch distillation (MBD) to maintain the discrimination of the new network with respect to the old features by aligning the invariant distribution knowledge on the old classes. 
  
  Specifically, we insert a residual adapter in each block of the old feature extractor to map the old representation to a high-dimensional embedding space, forcing the optimization flow to only pass through the expanding branches unrelated to the old class. After the optimization, we adopt the structural reparameterization technique to fuse the old and new features and map them back to the initial space losslessly. Furthermore, to reduce the confusion between the newly incremental classes and the original classes, we add a prototype selection mechanism (PSM) during the distillation process. The normalized cosine is first used to measure the similarity between the new representation and the old prototype. Then samples similar to the old classes are used for distillation, maintaining the old knowledge with a soft label that retains the old class statistical information, while those samples dissimilar to the old classes are used for new class training. This mechanism improves the performance of forward transfer and mitigates the lack of joint optimization to some extent. Our main contributions are as follows:
  
  1) A self-sustaining representation expansion scheme is proposed for non-exemplar incremental learning, in which a cyclically expanding optimization is accomplished by a dynamic structure reorganization strategy, resulting in a structure-invariant representation.
  
  2) A prototype selection mechanism is proposed, which combinatorially co-uses the preserved invariant knowledge and the incoming new supervision to reduce the feature confusion among the similar classes.
  
  3) Extensive experiments are performed on benchmarks including CIFAR-100, TinyImageNet and ImageNet-Subset, and the results demonstrate the superiority of our method over the state of the art.

\section{Related Work}
\label{sec:related work}
\subsection{Incremental Learning}
\label{related:incremental}
As deep learning research advances, there is a growing demand for continual learning of neural networks, which requires the network to learn new tasks without forgetting the old knowledge to achieve the stability-plasticity trade-off. 
CIL \cite{Ven2019ThreeSF, zhu2021self} is the most difficult scenery in continual learning and has received more attention recently. Current methods can broadly divided into the following three classes. Regularization-based methods \cite{kirkpatrick2017overcoming, zenke2017continual} estimate the importance of the network parameters learned in the past tasks and constrain their optimization accordingly. Rehearsal-based methods \cite{castro2018end, wu2019large, liu2020mnemonics, douillard2020podnet, Hu2021DistillingCE} preserve exemplars of fixed memory size to maintain the distribution of old classes in the incremental phases, and adopt the distillation skills to retain the discriminative features of the old task. \cite{hou2019learning} incorporates three components, cosine normalization, less-forget constraint, and inter-class separation, to address the imbalance between previous and new data. 
The techniques on exemplar and distillation in rehearsal-based methods are widely used in class-incremental learning. Structure-based methods \cite{Rusu2016ProgressiveNN, Hung2019CompactingPA} select and expand different sub-network structures involved in the optimization process of the incremental tasks. \cite{rajasegaran2019random} progressively chooses optimal paths for the new tasks while encouraging parameter sharing, which promotes the forward knowledge transfer. \cite{yan2021dynamically} freezes the previously learned representation and augment it with additional feature dimensions from a new mask-based feature extractor. 
The structure-based methods are often mixed with other techniques such as exemplar and distillation, and have achieved good results.

Recently, some works \cite{yu2020semantic, zhu2021prototype, yin2020dreaming} focus on a challenging but practical non-exemplar class-incremental learning problem, where no past data can be stored due to equipment limits or privacy security. \cite{yu2020semantic} estimates the semantic drift of incremental features and compensates the prototypes in each test phase. \cite{zhu2021prototype} adopts prototype augmentation to maintain the decision boundary of previous tasks, and employ self-supervised learning to learn more transferable features for future tasks. We follow their NECIL settings. However, different from their work considering generalizable features and augmented prototypes, we mainly consider the adjustment for joint representation learning and distillation process in the absence of exemplars.

\subsection{Residual Block}

Residual block has been widely used in convolutional neural network as the basic structure of ResNet \cite{he2016deep}, which improves the network depth and prevents vanishing gradient. Further improvements have been investigated for superior dynamic performance \cite{li2021revisiting} and inference efficiency \cite{Guo2020ExpandNetsLO, Ding2019ACNetST, ding2021repvgg} recently. In domain adaptation, residual adapter \cite{rebuffi2017learning, Rebuffi2018EfficientPO, Li2021ImprovingTA} is proposed to learn style information related to new domains, thus improving the overall generalization performance of the network. In these efforts, residual block is used to improve joint optimization performance or statiscal domain information. Instead, we consider dynamically incremental residual blocks to learn new knowledge efficiently while maintaining old features.



  
\section{Problem Description}
The NECIL problem is defined as follows. Here we denote X, Y and Z as the training set, the label set and the test set, respectively. Our task is to train the model from a continuous data stream, \textit{i.e.,} training sets $X^{1},  X^{2}, \cdots X^{n}$, where samples of a set $X^{i}$ ($1\leq i\leq n$) are from the label set $Y^{i}$, and $n$ represents the incremental phase. It should be mentioned that all the incremental classes are disjoint, that is, $Y^{i}\cap Y^{j}=\varnothing (i\neq j)$. Except that there are sufficient samples in the current phase $X^{i}$, no old samples are available in memory for old classes. To measure the performance of models in NECIL task, we calculate the classification accuracy on the test set $Z^{i}$ at each phase $i$. Different from the training set, the classes of the test set $Z^{i}$ are from all the seen label sets $Y^{1}\bigcup Y^{1}\cdots \bigcup Y^{i}$.  

\begin{figure*}[t]
    \centering
    \begin{overpic}[width=0.91\linewidth]{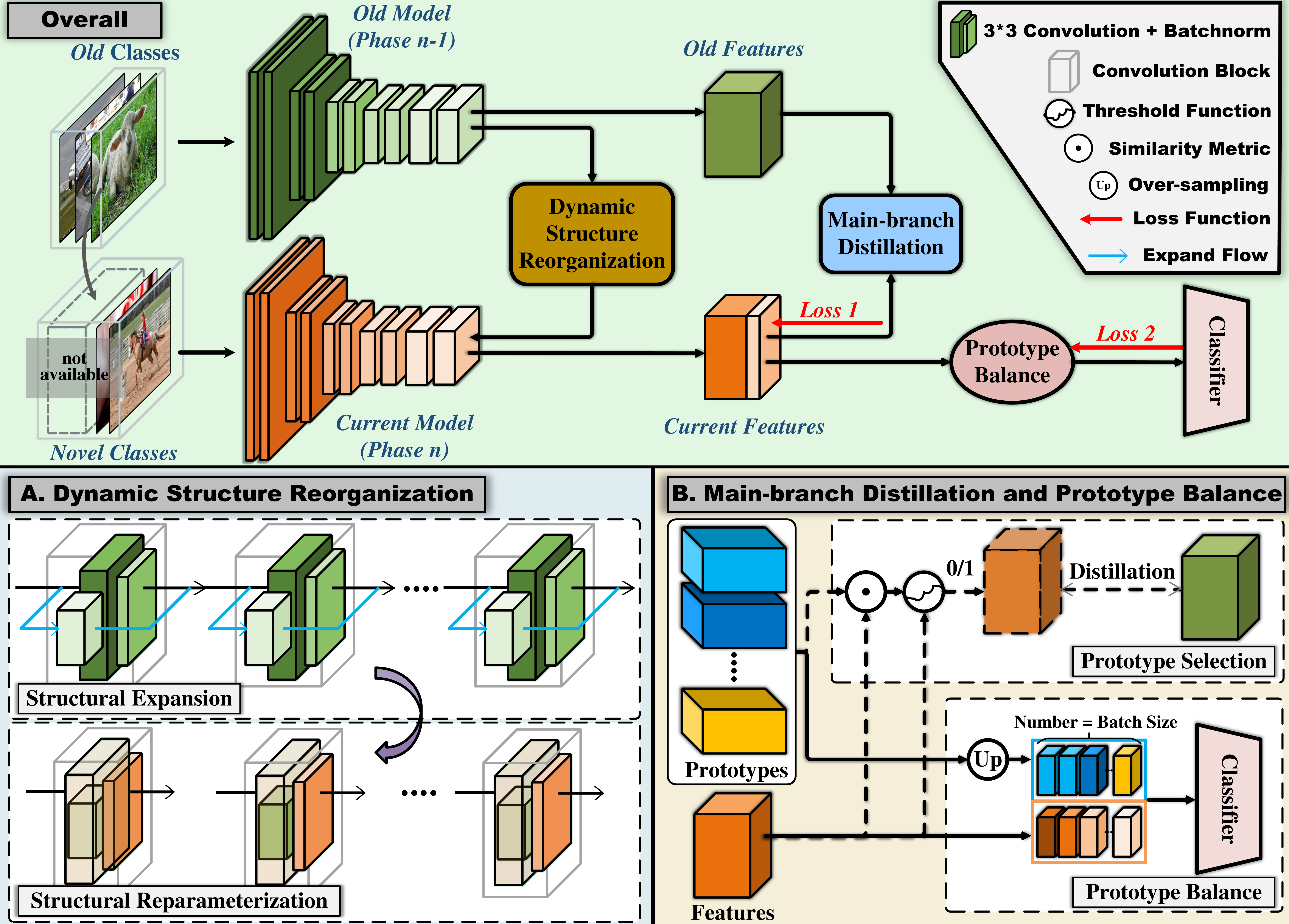}
    \put(29,22.1){\textbf{\small{(Eq. \ref{eqa:exapnsion})}}}
    \put(29,7.1){\textbf{\small{(Eq. \ref{eqa:repa})}}}
    \end{overpic}
  \caption{Our proposed self-sustaining representation expansion scheme for NECIL: (a) overview 
  of our scheme, (b) dynamic structure reorganization, and (c) main-branch distillation and prototype balance. The source code will be made available to the public.} 
  \label{fig:model}
\end{figure*}

\section{Methodology}
 First of all, we demonstrate the paradigms of standard CIL and how we adapt it to the NECIL setting as the baseline. Then we analyze the optimization flow of the overall pipeline and explain why it doesn't work well. Finally, two proposed core components dynamic structure reorganization and prototype selection are introduced.  

\subsection{Standard NECIL Paradigm}
\label{sec:baseline}
\cite{rebuffi2017icarl} first proposed a practical strategy for decoupling representation and classifiers learning in the CIL setting, which is followed by most subsequent work. The three main components are representation learning using knowledge distillation and prototype rehearsal, prioritized exemplar selection, and classification by a balance calibration.

\textbf{Incremental Representation Learning.}
As the exemplar cannot be saved in the NECIL setting, the representation learning will be slightly different, mainly in terms of cross-entropy loss and distillation loss. At the first phase, a standard classification model $f_{\theta }^{1}$ consisting of the feature extractor $f_{e}^{1}$ and classifier $g_{c}^{1}$ is optimized under the full supervision, \textit{i.e.}, $X^{1}$ and $Y^{1}$. At the incremental phase ($n>1$), the input of the current model is only the predicted images Q from $X^{n}$ without old samples. A base feature extractor $f_{e}^{n}$ such as VGG \cite{Simonyan2015VeryDC} or ResNet \cite{he2016deep} parameterized by $\theta_{e}^{n}$ is utilized to learn the corresponding representation:
\begin{equation}
   r_{q}^{n} = f_{e}^{n}(Q;\theta_{e}^{n}).
\end{equation}
Then, to learn the discriminative features among the novel classes, the obtained representation is optimized under the supervision of the class label y$_{q}^{n}$ from $Y^{n}$. We adopt the fully connected layer as the classifier $g_{c}^{n}$ to map the representation to the label space, 
\begin{align}
  s_{q}^{n} = g_{c}^{n}(r_{q}^{n}; \theta _{c}^{n}), L_{ce} = F_{ce}(s_{q}^{n}, y_{q}^{n}),
  \label{eq1}
\end{align}

$F_{ce}$ represents the standard cross-entropy loss. Finally, to maintain the useful information learned with the old classes, the knowledge distillation is used to measure the similarity between the obtained representation and that of the previous model $f_{e}^{n-1}$,
\begin{align}
  r_{q}^{n-1} = f_{e}^{n-1}(Q;\theta_{e}^{n-1}), L_{kd} = F_{kd}(r_{q}^{n}, r_{q}^{n-1}),
  \label{eq2}
\end{align}
$F_{kd}$ represents Euclidean distance the same as \cite{zhu2021prototype}.
  
\textbf{Incremental Classifier Calibration.}
A common way to overcome the imbalance between the exemplars and new samples in CIL is to under-sample a balanced subset for fine-tuning. As there is no exemplars in NECIL, we memorize one prototype in the deep feature space for each class, which is consistent with PASS \cite{zhu2021prototype}. Different from PASS augmenting the prototypes via Gaussian noise, we choose to over-sample (\textit{i.e.}, $Up_{B}$) prototypes to the batch size (\textit{i.e.}, B), achieving the calibration of the classifier, which is the simplest way adopted in the long-tail recognition~\cite{chawla2002smote},
\begin{align}
  p_{B} = Up_{B}(Prototype), L_{proto} = F_{ce}(p_{B}, y_{B}),
\end{align}
$y_{B}$ is the over-sampled label set of the initial prototypes. The final loss for current model is their addition:
\begin{equation}
L = L_{ce} + \lambda L_{kd} + \gamma L_{proto},
\end{equation}
$\lambda$ and $\gamma$ are loss weights, and we set them to 10.
  
\subsection{Optimization}
Different from the previous work focusing on the effect on the classifier, this paper tries to analyze the representation. In the CIL, 
Equation \ref{eq1} can be turned into two parts:
\begin{equation}
  L_{ce} = F_{ce}(s_{q}^{n}, y_{q}^{n}) + F_{ce}(s_{e}^{n}, y_{e}^{n}),
\end{equation}
$s_{e}^{n}$ represents the saved exemplars, whose number is much lower than that of $s_{q}^{n}$. While this imbalance can bias the optimization process towards features that are more discriminative for the new class, the added distillation in Equation \ref{eq2} can alleviate this problem, 
\begin{equation}
  L_{kd} = F_{kd}(r_{q}^{n}, r_{q}^{n-1}) + F_{kd}(r_{e}^{n}, r_{e}^{n-1}).
\end{equation}
$r_{e}^{n}$ represents the representation of exemplars.
In this case, the features that are significant for the old and new classes will be maintained. However, note that 
there is no exemplars involved in the above NECIL setting. It means that the joint optimization on the old and new class representations completely collapses into feature optimization that is relevant only to incremental classes. What is reflected in the first part is that the cross-entropy loss will only focus on the features that facilitate the recognition of the new class, while in the second part, it will focus on the maintenance of the features related to the new class, which both accelerate the forgetting of the representative features of the old class. Suppose the forgetting rate (Fr) of distillation part in the initial phase is $\alpha $. Note that the distillation loss is based on the overall representation of the previous phase, and the error will accumulate exponentially with the phase, that is, $Fr_{n} \ge \alpha ^{n-1}$.
Therefore, it is necessary to correct this error from the representational level.
  
\subsection{Self-Sustaining Representation Expansion}
\textbf{Dynamic Structure Reorganization.} To retain the representation of the old class and guarantee the unbiased training of the new class, we propose a dynamic structure reorganization strategy. In general, as shown in Fig. \ref{fig:model}, we firstly adopt the structural expansion to add the side branch to the current model by block for the optimization of new classes. Specifically, we insert a residual adapter to each convolution block of the fixed feature extractor from previous phase. The optimized flow propagates only through the adapter, updating the most discriminating position while maintaining the old features,   
\begin{equation}
\label{eqa:exapnsion}
\begin{aligned}
  f_{e}^{n}(Q;\theta_{e}^{n}) = F_{transform}&(f_{e}^{n-1}(Q;\theta_{e}^{n-1})) \\
  &=f_{e}^{n-1}(Q;\hat{\theta}_{e}^{n-1}\oplus \Delta \theta_{e}^{n}),
\end{aligned}
\end{equation}
where $\hat{}$ represents the fixed parameters, and $\oplus$ represents the structural expansion operation. After training, we use the structural reparameterization \cite{ding2021repvgg} to integrate the side-branch information into the main branch losslessly, ensuring that the number of network parameters does not increase at the end of each phase.
Specifically, the parameters in the residual structures are fused with the parameters of the original convolution kernel and BatchNorm \cite{ioffe2015batch} through the zero-padding operation and linear transformation, 
and finally the adapters are removed to keep the network structure unchanged for the next update, 
\begin{equation}
\label{eqa:repa}
\begin{aligned}
  f_{e}^{n-1}(Q;\hat{\theta}_{e}^{n-1}&\oplus \Delta \theta_{e}^{n}) \\
  &=f_{e}^{n}(Q;{\theta_{e}^{n}}' \oplus 0) = f_{e}^{n}(Q;{\theta_{e}^{n}}').
\end{aligned}
\end{equation}
  
\textbf{Prototype Selection.} While new features are learned based on the old structure, the old class features are maintained in coordination with the main-branch distillation. To reduce the feature confusion in the distillation part, we adopt a prototype selection mechanism based on the expandable embedding space. In general, based on the similarity between the representation of new samples and the old prototypes, dissimilar 
samples are involved in the update of residual adapter to learn new features, and similar samples are involved in the distillation to retain the old discriminative features maintained in the main branch at previous phase. Specifically, after mapping all new samples to the learned embedding space, we compute the normalized cosine scores $Si$ between them and all prototypes, 
\begin{equation}
  Si = Cosine(N(r_{q}^{n}), Nor(Prototype)),
\end{equation}
Nor represents the normalization operation. We then set a threshold value, and attach a mask to the corresponding position of its distillation loss ($Mask_{kd}$) if greater than the threshold $\sigma$, and add a mask to the corresponding part of its cross-entropy loss ($Mask_{ce}$) if less than the threshold. Finally, the two loses are summed with the prototype balance loss as the final optimization function for the new phase.
\begin{align}
  &L = Mask_{ce}(L_{ce}) + \lambda Mask_{kd}(L_{kd}) + \gamma L_{proto}.
\end{align}

\section{Experiments}
\subsection{Dataset and Settings}
\textbf{Dataset.} To evaluate the performance of our proposed method, we conduct comprehensive experiments on 
three datasets CIFAR-100 \cite{Krizhevsky2009LearningML}, TinyImageNet \cite{le2015tiny} and ImageNet-Subset. CIFAR-100 contains 60000 images of 32 $\times$ 32 size from 100 classes, 
and each class includes 500 training images and 100 test images. TinyImageNet contains 200 classes, and each class contains 
500 training images, 50 validation images and 50 test images. It provides more phases and incremental classes to 
compare the sensitivity of different methods. ImageNet-Subset is a 100-class subset of ImageNet-1k \cite{deng2009imagenet}, which is much larger. For the order and division of all dataset classes in our experiments, we followed exactly the settings in \cite{zhu2021prototype}.

  
\textbf{Setting.} As adopted in \cite{zhu2021prototype}, we use ResNet-18 as the backbone network. The difference is that 
we use standard supervised training for the whole optimization process instead of involving self-supervised learning. 
For a fair comparison, we achieve the same accuracy as \cite{zhu2021prototype} at the first phase for all datasets. We use an Adam optimizer, in which the initial learning rate is set to 0.001 and the attenuation rate is set to 0.0005. The model stops training after 100 epochs, and batch size is set to 128.

  \begin{table}[t]
\small
    \renewcommand{\arraystretch}{1.}
  \renewcommand{\tabcolsep}{6.pt}
  \centering
  \begin{tabular}{ccc||ccc}
    \hline
    \Xhline{2.\arrayrulewidth}
  \multirow{2}{*}{DSR} & \multirow{2}{*}{MBD} & \multirow{2}{*}{PSM} & \multicolumn{3}{c}{CIFAR-100}    \\
  \cline{4-6} 
                            &                               & & 5 phases & 10 phases & 20 phases \\ 
    \hline
    \Xhline{2.\arrayrulewidth}
                            &                               & & 61.11    & 57.08     & 51.04     \\ 
  $\surd$                   &                               & & 64.86    & 63.25     & 54.09     \\ 
                            & $\surd$                       & & 62.70    & 62.60     & 58.57     \\ 
  $\surd$                   & $\surd$                       & & 65.10    & 63.87     & 60.60     \\ 
\rowcolor{mygray}
  $\surd$                   & $\surd$            & $\surd$  & 65.88    & 64.69     & 61.61     \\ 
    \hline
    \Xhline{2.\arrayrulewidth}
  \end{tabular}
  \caption{Ablation study of our method on CIFAR-100.}
  \label{ablation}
\end{table}

\begin{table}[t]
\small
    \renewcommand{\arraystretch}{1.}
  \renewcommand{\tabcolsep}{9.pt}
  \centering
  \begin{tabular}{c|ccc}
  \hline
    \Xhline{2.\arrayrulewidth}
               & \multicolumn{3}{c}{CIFAR-100}                                                   \\ 
               \hline
  Method       & \multicolumn{1}{c|}{5 phases} & \multicolumn{1}{c|}{10 phases} & 20 phases \\ 
  \hline
    \Xhline{2.\arrayrulewidth}
  3$\times $3 conv      &  \multicolumn{1}{c|}{64.28}           & \multicolumn{1}{c|}{63.47}            &60.81             \\ \cline{1-1}
  1$\times $1 conv + bn & \multicolumn{1}{c|}{65.88}           & \multicolumn{1}{c|}{64.84}            &60.72             \\ \cline{1-1}
  \rowcolor{mygray}
  1$\times $1 conv      & \multicolumn{1}{c|}{65.87}           & \multicolumn{1}{c|}{65.12}            &61.60             \\ 
  \hline
    \Xhline{2.\arrayrulewidth}
  \end{tabular}
  \caption{Performance under different expanding structures.}
  \label{an:adapter}
  \end{table}
  
\textbf{Evaluation Metrics.}
Following \cite{zhu2021prototype}, we report average incremental accuracy and average forgetting, and our performance 
is evaluated on three different runs. Average incremental accuracy is 
computed as the average accuracy of all the incremental phases (including the first phase), which  
compares the overall incremental performance of different methods fairly. Average forgetting is computed as 
the average forgetting of different tasks throughout the incremental process, which directly measures the 
ability of different methods to resist catastrophic forgetting.

\subsection{Ablation Study}
To prove the effectiveness of our proposed method, we conduct several ablation experiments on CIFAR-100. The performance of our scheme is mainly attributed to two prominent components: the dynamic 
structure reorganization strategy (DSR) and the main-branch distillation (MBD). To clarify the function of DSR, we replace the dynamic representation with the structurally invariant representation, which is adopted in most CIL methods \cite{zhu2021prototype, douillard2020podnet}. To clarify the 
function of MBD, we replace the distillation process with the one that interacts with the continuously optimized representation.
As can be seen in Table \ref{ablation}, the dynamic representation and main-branch distillation separately bring a 4.3$\%$ and 4.8$\%$ improvement in overall performance. It demonstrates that the two parts are far more useful than standard representation and common distillation respectively in NECIL. It is worth noting that the former plays a greater role when there are fewer incremental phases (\textit{i.e.}, 5 and 10 phases), while the latter shines more brightly when there are more incremental phases (\textit{i.e.}, 20 phases). It demonstrates that keeping the old class features helps to improve the overall performance of incremental learning in the short term. However, as analyzed in the introduction, if the distilled network keep decaying or fixed, the errors will accumulate as the incremental phase increases. At this point, how to reasonably correct the distillation loss is the key to ensure the long-term effect.

     \begin{figure}[t]
 \centering
  \begin{overpic}[width=0.9\linewidth]{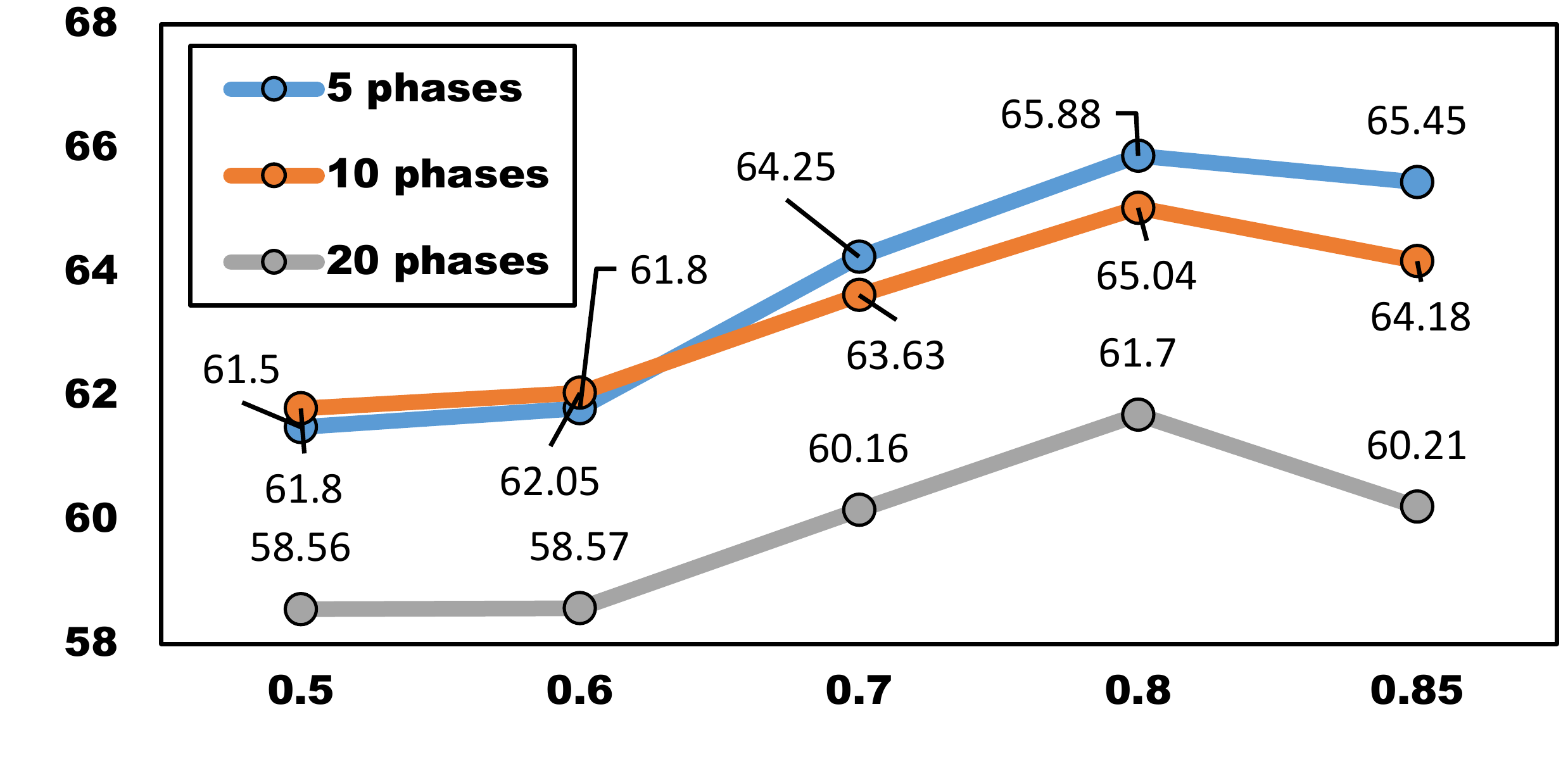}
  \put(-1,19){\rotatebox{90}{\textbf{\small{Accuracy}}}}
  \put(46,-1.){\textbf{\small{Threshold}}}
    \end{overpic}
     \caption{Illustration of the role of the selection mechanism.}
  \label{fig:an2}
\end{figure}

\begin{figure}[t]
\centering
  \begin{overpic}[width=0.99\linewidth]{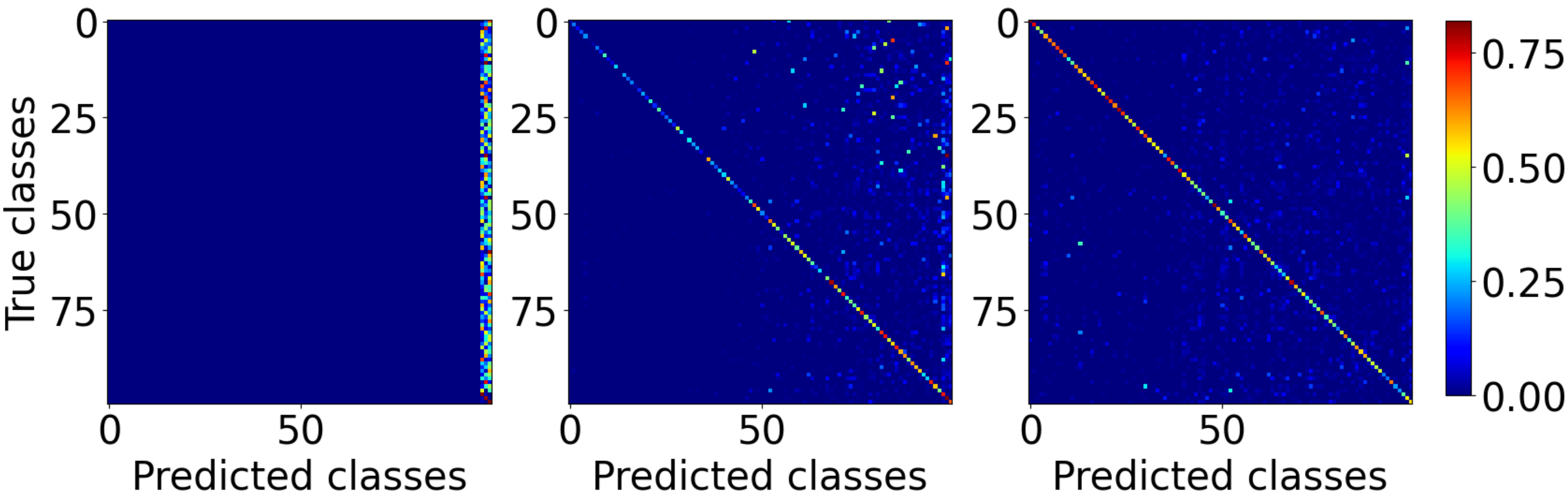}
  \put(6,-3.5){\textbf{\small{(a) Finetuning}}}
  \put(40,-3.5){\textbf{\small{(b) iCaRL}}}
  \put(71,-3.5){\textbf{\small{(c) Ours}}}
    \end{overpic}
     \caption{Confusion matrices of different methods on CIFAR-100.}
  \label{fig:confusion}
\end{figure}

  \begin{figure*}[h]
\centering
    \begin{overpic}[width=0.9\linewidth]{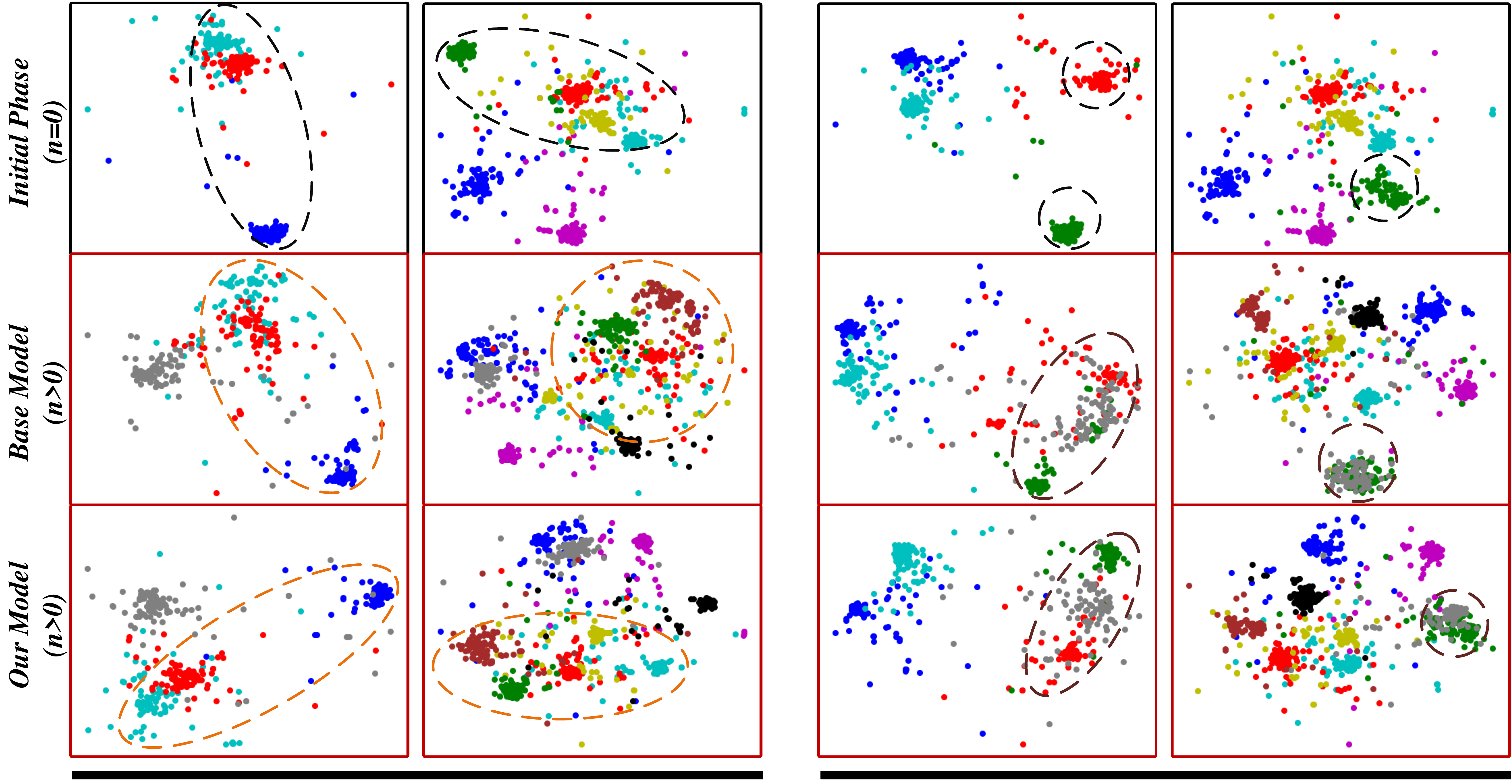}
  \put(26.5,-1.8){\textbf{\small{(a)}}}
  \put(76,-1.8){\textbf{\small{(b)}}}
  \end{overpic}
       \caption{Effect of our scheme on the representation. (a) DSR maintains the discriminative features and inter-relations of old classes, thus enhancing the clustering and separation of the distribution of old classes. (b) MBD results in a better distinction between similar classes.}
    \label{fig:visual}
  \end{figure*}
  
  \begin{table*}[ht]
\small
  \renewcommand{\arraystretch}{1.}
  \renewcommand{\tabcolsep}{5.pt}
    \centering
    \begin{tabular}{l|l||ccc|ccc|c}
    \hline
    \Xhline{2.\arrayrulewidth}
    \multicolumn{2}{c||}{} & \multicolumn{3}{c|}{\textbf{CIFAR-100}} & \multicolumn{3}{c|}{\textbf{TinyImageNet}} & \multicolumn{1}{c}{\textbf{ImageNet-Subset}} \\ \cline{3-5} \cline{6-8} \cline{9-9}
    \multicolumn{2}{c||}{\multirow{-2}{*}{\textbf{Methods}}} & \textit{P=5} & \textit{P=10} & \textit{P=20} & \textit{P=5} & \textit{P=10} & \textit{P=20} & \textit{P=10} \\ 
    \hline
    \Xhline{2.\arrayrulewidth}
      & iCaRL-CNN$^{\ast }$   & 51.07  & 48.66 & 44.43 & 34.64 & 31.15 & 27.90 & 50.53 \\
      & iCaRL-NCM$^{\ast }$ \cite{rebuffi2017icarl}  & 58.56 & 54.19 & 50.51 & 45.86 & 43.29 & 38.04 & 60.79  \\ 
      & EEIL$^{\ast }$ \cite{castro2018end} & 60.37 & 56.05 & 52.34 & 47.12 & 45.01 & 40.50 & 63.34  \\
    \multirow{-4}{*}{\rotatebox{90}{\textit{(1) E=20}}} & UCIR$^\ast $ \cite{hou2019learning} & 63.78 & 62.39 & 59.07 & 49.15 & 48.52 & 42.83 & 66.16   \\ 
    \hline
      & EWC$^{\ast }$ \cite{kirkpatrick2017overcoming} & 24.48 & 21.20 & 15.89 & 18.80 & 15.77 & 12.39 & 20.40 \\ 
      & LwF$\_$MC$^{\ast }$ \cite{rebuffi2017icarl}  & 45.93 & 27.43 & 20.07 & 29.12 & 23.10 & 17.43 & 31.18  \\
      & MUC$^{\ast }$ \cite{2020More} & 49.42 & 30.19 & 21.27 & 32.58 & 26.61 & 21.95 & 35.07 \\ 
      & SDC \cite{yu2020semantic} & 56.77 & 57.00 & 58.90 & - & - & - & 61.12 \\ 
      & PASS \cite{zhu2021prototype} & 63.47 & 61.84 & 58.09 & 49.55 & 47.29 & 42.07 & 61.80  \\
      \rowcolor{mygray}
    \multirow{-6}{*}{\rotatebox{90}{\textit{(2) E=0}}} & Ours & \textbf{65.88}\footnotesize{\color{red} \textbf{+2.41}} & \textbf{65.04}\footnotesize{\color{red} \textbf{+3.20}} & \textbf{61.70}\footnotesize{\color{red} \textbf{+2.80}} & \textbf{50.39}\footnotesize{\color{red} \textbf{+0.84}} & \textbf{48.93}\footnotesize{\color{red} \textbf{+1.64}} & \textbf{48.17}\footnotesize{\color{red} \textbf{+6.10}}  & \textbf{67.69}\footnotesize{\color{red} \textbf{+5.89}} \\ 
    \hline
    \Xhline{2.\arrayrulewidth}
    \end{tabular}
    \vspace{-0.1in}
    \caption{Comparisons of the average incremental accuracy (\%) with other methods on 
    CIFAR-100, TinyImageNet, and ImageNet-Subset. P represents the number of phases and E represents 
    the number of exemplars. Models with an asterisk $^{\ast }$ represent the reproduced 
    results in \cite{zhu2021prototype}. The red footnotes in the last row represent the 
    relative improvement compared with the results of SOTA.}
    \label{tab:sota}
    \vspace{-0.15in}
  \end{table*}
  
\subsection{Analysis}
\textbf{The impact of the adapter structure.}
To explore the impact of the structure of residual adapter on expandable representation during training, we design the 
following experiments. We adopt three different convolution blocks to the residual part: 1$\times $1 
convolution only, 3$\times $3 convolution only and the combination of 1$\times $1 convolution and BatchNorm. As 
shown in Table \ref{an:adapter}, the results of 1$\times $1 convolution and the combination are similar, and that of 3$\times $3 
convolution is one point lower. It suggests that the 1$\times $1 convolution structure is good enough to learn the representation of the new class without needing more parameters. 

  \vspace{3mm}
  \begin{figure*}[t]
\centering
    \begin{overpic}[width=0.95\linewidth]{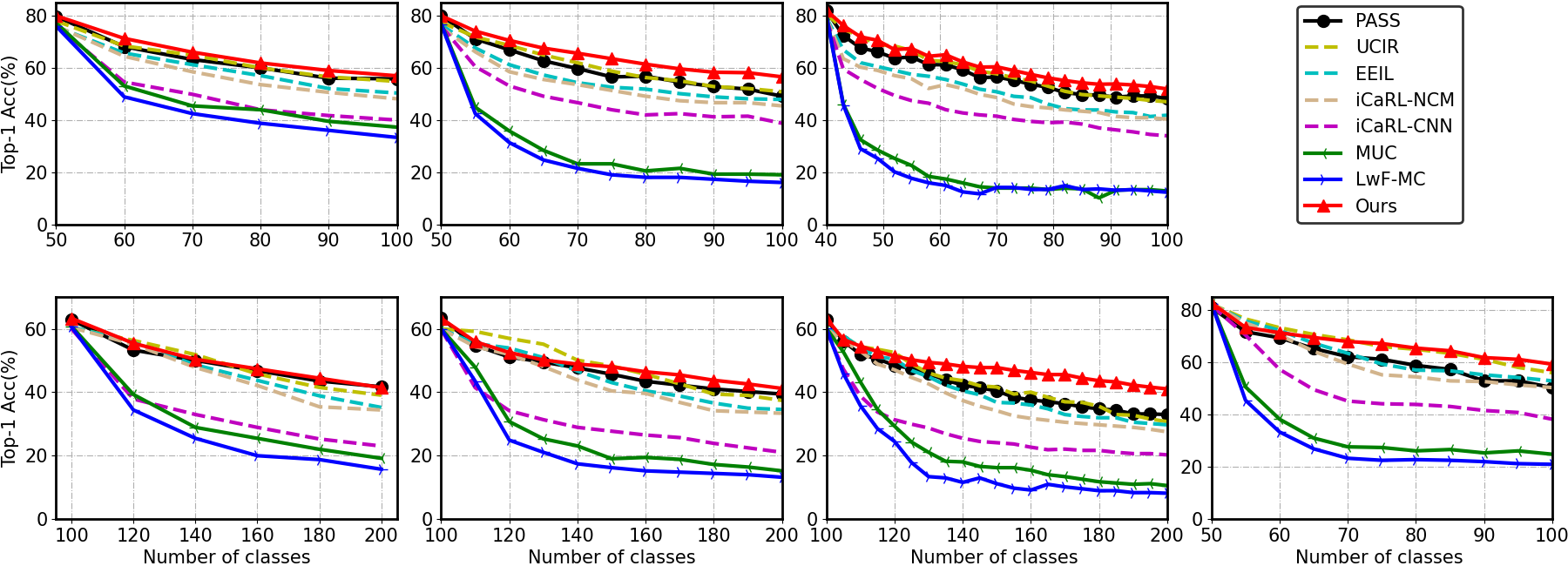}
  \put(3,18.4){\textbf{\small{(a) 5 phases CIFAR-100}}}
  \put(28,18.4){\textbf{\small{(b) 10 phases CIFAR-100}}}
  \put(53,18.4){\textbf{\small{(c) 20 phases CIFAR-100}}}
  \put(2,-1.8){\textbf{\small{(d) 5 phases TinyImagNet}}}
  \put(27,-1.8){\textbf{\small{(e) 10 phases TinyImagNet}}}
  \put(52,-1.8){\textbf{\small{(f) 20 phases TinyImagNet}}}
  \put(76,-1.8){\textbf{\small{(g) 10 phases ImagNet-Subset}}}
  \end{overpic}
       \caption{Classification accuracy on CIFAR-100, TinyImageNet and ImageNet-Subset, which contains the complete curves.}
    \label{fig:curves}
  \end{figure*}
  
\begin{figure}[t]
\footnotesize
\centering
  \begin{overpic}[width=0.99\linewidth]{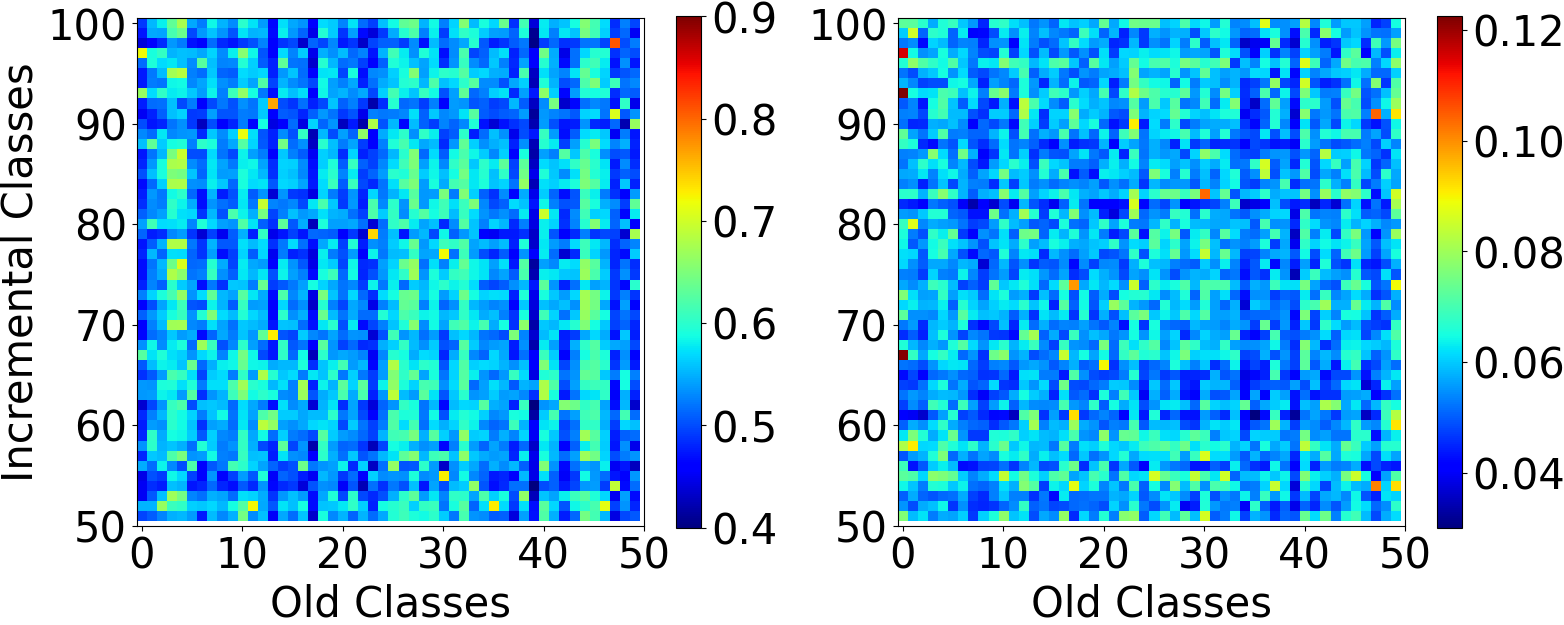}
  \put(8,-3.){\textbf{(a) Mean of similarity}}
  \put(50,-3.){\textbf{(b) Standard deviation of similarity}}
    \end{overpic}
  \caption{Statistics of similarity on the incremental samples.}
  \label{fig:visual2}
\end{figure}

\textbf{The impact of the threshold in prototype selection.}
To verify the role of the PSM, we conduct data statistics on the incremental samples. As shown in Fig. \ref{fig:visual2}, the new classes have a large difference in similarity. And the intra-class fluctuations are also large, so different classes and samples involved in the optimization process will bring different changes. Therefore it is important to reasonably place them in the two potentially conflicting processes of old feature distillation and new feature learning. To demonstrate the sensitivity of the threshold on the distillation effect, we plot its fluctuation curve. As shown in Fig. \ref{fig:an2}, all curves rise to a peak at a threshold of 0.8, then gradually fall and lose distillation effect. It suggests that in the absence of the exemplars, fine-grained optimization of the new samples can better maintain the old features and learn the new features.

\textbf{Classification accuracy of old and novel classes.}
To evaluate performance of both old and new classes during training, we compare their accuracy at each phase. As shown in Fig. \ref{fig:confusion}, our method achieves similar performance between the old and new classes without favoring one side due to overfitting, which is a prerequisite for a good incremental learning system.

\subsection{Visualization}
To better demonstrate the role of DSR and MBD during optimization, we show the visualization results with t-SNE \cite{maaten2008visualizing}. As shown in Fig. \ref{fig:visual} (a), although the old classes have slightly changed in the representation after multi-phase optimization, their discrimination and relative relationship almost do not decline with our DSR. As shown in Fig. \ref{fig:visual} (b), newly incremental classes are easily confused with some of the old classes. Owing to our MBD, the optimized features are promoted to differentiate from the old class, thus improving the seperation of novel clusters.
  


\subsection{Comparison with SOTA}
To better assess the overall performance of our scheme, we compare it to the SOTA of NECIL (EWC$^{\ast }$, LwF$\_$MC$^{\ast }$, MUC$^{\ast }$, SDC and PASS) and some classical methods of exemplar-based CIL (iCARL$^{\ast }$, EEIL$^{\ast }$ and UCIR$^{\ast }$).

\textbf{Average accuracy and average forgetting.}
As shown in Table \ref{tab:sota}, compared to the SOTA of non-exemplar methods (E=0), our method achieves average improvement of 3, 3 and 6 points on CIFAR-100, TinyImageNet and ImageNet-Subset, respectively. The performance of our method is comparable to the classical exemplar-based methods (E=20), which shows that our scheme further reduces the impact of exemplars on CIL models. To provide further insight into the behaviors of different methods, we compare their average forgetting of all phases. As shown in Table \ref{tab:forget}, our method achieves much lower average forgetting, resisting catastrophic forgetting well in the absence of exemplars.


\begin{table}[t]
\small
\renewcommand{\arraystretch}{1.}
  \renewcommand{\tabcolsep}{4.pt}
\centering
\begin{tabular}{l||ccc|ccc}
    \hline
    \Xhline{2.\arrayrulewidth}
          & \multicolumn{3}{c|}{CIFAR-100}                                             & \multicolumn{3}{c}{TinyImageNet}                                           \\ \hline
Method    & \multicolumn{1}{c|}{5} & \multicolumn{1}{c|}{10} & \multicolumn{1}{c|}{20} & \multicolumn{1}{c|}{5} & \multicolumn{1}{c|}{10 } & \multicolumn{1}{c}{20} \\ 
    \hline
    \Xhline{2.\arrayrulewidth}
iCaRL-CNN & \multicolumn{1}{c|}{42.13}    & \multicolumn{1}{c|}{45.69}     & 43.54     & \multicolumn{1}{c|}{36.89}    & \multicolumn{1}{c|}{36.70}     & 45.12     \\
iCaRL-NCM & \multicolumn{1}{c|}{24.90}    & \multicolumn{1}{c|}{28.32}     & 35.53     & \multicolumn{1}{c|}{27.15}    & \multicolumn{1}{c|}{28.89}     & 37.40     \\
EEIL      & \multicolumn{1}{c|}{23.36}    & \multicolumn{1}{c|}{26.65}     & 32.40     & \multicolumn{1}{c|}{25.56}    & \multicolumn{1}{c|}{25.91}     & 35.04     \\
UCIR      & \multicolumn{1}{c|}{21.00}    & \multicolumn{1}{c|}{25.12}     & 28.65     & \multicolumn{1}{c|}{20.61}    & \multicolumn{1}{c|}{22.25}     & 33.74     \\ \hline
LwF\_MC   & \multicolumn{1}{c|}{44.23}    & \multicolumn{1}{c|}{50.47}     & 55.46     & \multicolumn{1}{c|}{54.26}    & \multicolumn{1}{c|}{54.37}     & 63.54     \\
MUC       & \multicolumn{1}{c|}{40.28}    & \multicolumn{1}{c|}{47.56}     & 52.65     & \multicolumn{1}{c|}{51.46}    & \multicolumn{1}{c|}{50.21}     & 58.00     \\
PASS      & \multicolumn{1}{c|}{25.20}    & \multicolumn{1}{c|}{30.25}     & 30.61     & \multicolumn{1}{c|}{18.04}    & \multicolumn{1}{c|}{23.11}     & 30.55     \\
 \rowcolor{mygray}
Ours       & \multicolumn{1}{c|}{\textbf{18.37}}    & \multicolumn{1}{c|}{\textbf{19.48}}     & \textbf{19.00}     & \multicolumn{1}{c|}{\textbf{9.17}}     & \multicolumn{1}{c|}{\textbf{14.06}}     & \textbf{14.20}     \\ 
    \hline
    \Xhline{2.\arrayrulewidth}
\end{tabular}
\caption{Results of average forgetting on 5, 10 and 20 phases.}
\label{tab:forget}
\end{table}

\textbf{Trend of accuracy.}
To analyze the trend of different methods, we show the detailed accuracy curves on three datasets. As shown in Fig. \ref{fig:curves}, our method is superior at almost all phases, striking a better stability-plasticity balance. It can be seen that the difficulty increases as the number of incremental phases (P) increases. In this process, the advantage of our method are even expanding, such as in TinyImageNet. Whether in the smaller CIFAR-100 or the larger ImageNet-Subset dataset, our method has a notable advantage, demonstrating its robustness.

\section{Conclusion and Discussion}
In this paper, a novel self-sustaining representation expansion scheme is presented for the NECIL task. A dynamic structure reorganization strategy is first proposed to optimize the newly incremental features in a side branch while maintaining the old feature distribution from the structurally expanded direction, and then the distillation process is arranged in the main branch. In particular, a prototype selection mechanism is integrated into the joint training to enhance the distinction between the old and new classes. Experimental results show that our method is superior in both performance and adaptability to the state-of-the-art methods, especially in the multi-phase process.

\textbf{Acknowledgments.} Supported by National Key R\&D Program of China under Grant 2020AAA0105701, National Natural Science Foundation of China (NSFC) under Grants 61872327 and Major Special Science and Technology Project of Anhui (No. 012223665049)






{\small
\bibliographystyle{ieee_fullname}
\bibliography{egbib}

\begin{thebibliography}{10}\itemsep=-1pt

\bibitem{castro2018end}
Francisco~M Castro, Manuel~J Mar{\'\i}n-Jim{\'e}nez, Nicol{\'a}s Guil, Cordelia
  Schmid, and Karteek Alahari.
\newblock End-to-end incremental learning.
\newblock In {\em Proceedings of the European conference on computer vision
  (ECCV)}, pages 233--248, 2018.

\bibitem{chawla2002smote}
Nitesh~V Chawla, Kevin~W Bowyer, Lawrence~O Hall, and W~Philip Kegelmeyer.
\newblock Smote: synthetic minority over-sampling technique.
\newblock {\em Journal of artificial intelligence research}, 16:321--357, 2002.

\bibitem{deng2009imagenet}
Jia Deng, Wei Dong, Richard Socher, Li-Jia Li, Kai Li, and Li Fei-Fei.
\newblock Imagenet: A large-scale hierarchical image database.
\newblock In {\em 2009 IEEE conference on computer vision and pattern
  recognition}, pages 248--255. Ieee, 2009.

\bibitem{Ding2019ACNetST}
Xiaohan Ding, Yuchen Guo, Guiguang Ding, and J. Han.
\newblock Acnet: Strengthening the kernel skeletons for powerful cnn via
  asymmetric convolution blocks.
\newblock {\em 2019 IEEE/CVF International Conference on Computer Vision
  (ICCV)}, pages 1911--1920, 2019.

\bibitem{ding2021repvgg}
Xiaohan Ding, Xiangyu Zhang, Ningning Ma, Jungong Han, Guiguang Ding, and Jian
  Sun.
\newblock Repvgg: Making vgg-style convnets great again.
\newblock In {\em Proceedings of the IEEE/CVF Conference on Computer Vision and
  Pattern Recognition}, pages 13733--13742, 2021.

\bibitem{douillard2020podnet}
Arthur Douillard, Matthieu Cord, Charles Ollion, Thomas Robert, and Eduardo
  Valle.
\newblock Podnet: Pooled outputs distillation for small-tasks incremental
  learning.
\newblock In {\em Computer Vision--ECCV 2020: 16th European Conference,
  Glasgow, UK, August 23--28, 2020, Proceedings, Part XX 16}, pages 86--102.
  Springer, 2020.

\bibitem{french1999catastrophic}
Robert~M French.
\newblock Catastrophic forgetting in connectionist networks.
\newblock {\em Trends in cognitive sciences}, 3(4):128--135, 1999.

\bibitem{Guo2020ExpandNetsLO}
Shuxuan Guo, Jos{\'e}~Manuel {\'A}lvarez, and Mathieu Salzmann.
\newblock Expandnets: Linear over-parameterization to train compact
  convolutional networks.
\newblock {\em arXiv: Computer Vision and Pattern Recognition}, 2020.

\bibitem{he2016deep}
Kaiming He, Xiangyu Zhang, Shaoqing Ren, and Jian Sun.
\newblock Deep residual learning for image recognition.
\newblock In {\em Proceedings of the IEEE conference on computer vision and
  pattern recognition}, pages 770--778, 2016.

\bibitem{hou2019learning}
Saihui Hou, Xinyu Pan, Chen~Change Loy, Zilei Wang, and Dahua Lin.
\newblock Learning a unified classifier incrementally via rebalancing.
\newblock In {\em Proceedings of the IEEE/CVF Conference on Computer Vision and
  Pattern Recognition}, pages 831--839, 2019.

\bibitem{Hu2021DistillingCE}
Xinting Hu, Kaihua Tang, Chunyan Miao, Xiansheng Hua, and Hanwang Zhang.
\newblock Distilling causal effect of data in class-incremental learning.
\newblock {\em 2021 IEEE/CVF Conference on Computer Vision and Pattern
  Recognition (CVPR)}, pages 3956--3965, 2021.

\bibitem{Hung2019CompactingPA}
Steven C.~Y. Hung, Cheng-Hao Tu, Cheng-En Wu, Chien-Hung Chen, Yi-Ming Chan,
  and Chu-Song Chen.
\newblock Compacting, picking and growing for unforgetting continual learning.
\newblock {\em ArXiv}, abs/1910.06562, 2019.

\bibitem{ioffe2015batch}
Sergey Ioffe and Christian Szegedy.
\newblock Batch normalization: Accelerating deep network training by reducing
  internal covariate shift.
\newblock In {\em International conference on machine learning}, pages
  448--456. PMLR, 2015.

\bibitem{kirkpatrick2017overcoming}
James Kirkpatrick, Razvan Pascanu, Neil Rabinowitz, Joel Veness, Guillaume
  Desjardins, Andrei~A Rusu, Kieran Milan, John Quan, Tiago Ramalho, Agnieszka
  Grabska-Barwinska, et~al.
\newblock Overcoming catastrophic forgetting in neural networks.
\newblock {\em Proceedings of the national academy of sciences},
  114(13):3521--3526, 2017.

\bibitem{Krizhevsky2009LearningML}
A. Krizhevsky.
\newblock Learning multiple layers of features from tiny images.
\newblock 2009.

\bibitem{le2015tiny}
Ya Le and Xuan Yang.
\newblock Tiny imagenet visual recognition challenge.
\newblock {\em CS 231N}, 7(7):3, 2015.

\bibitem{Li2021ImprovingTA}
Wei-Hong Li, Xialei Liu, and Hakan Bilen.
\newblock Improving task adaptation for cross-domain few-shot learning.
\newblock {\em ArXiv}, abs/2107.00358, 2021.

\bibitem{li2021revisiting}
Yunsheng Li, Yinpeng Chen, Xiyang Dai, Mengchen Liu, Dongdong Chen, Ye Yu, Lu
  Yuan, Zicheng Liu, Mei Chen, and Nuno Vasconcelos.
\newblock Revisiting dynamic convolution via matrix decomposition.
\newblock {\em arXiv preprint arXiv:2103.08756}, 2021.

\bibitem{liu2021adaptive}
Yaoyao Liu, Bernt Schiele, and Qianru Sun.
\newblock Adaptive aggregation networks for class-incremental learning.
\newblock In {\em Proceedings of the IEEE/CVF Conference on Computer Vision and
  Pattern Recognition}, pages 2544--2553, 2021.

\bibitem{liu2020mnemonics}
Yaoyao Liu, Yuting Su, An-An Liu, Bernt Schiele, and Qianru Sun.
\newblock Mnemonics training: Multi-class incremental learning without
  forgetting.
\newblock In {\em Proceedings of the IEEE/CVF Conference on Computer Vision and
  Pattern Recognition}, pages 12245--12254, 2020.

\bibitem{maaten2008visualizing}
Laurens van~der Maaten and Geoffrey Hinton.
\newblock Visualizing data using t-sne.
\newblock {\em Journal of machine learning research}, 9(Nov):2579--2605, 2008.

\bibitem{rajasegaran2019random}
Jathushan Rajasegaran, Munawar Hayat, Salman Khan, Fahad~Shahbaz Khan, and Ling
  Shao.
\newblock Random path selection for incremental learning.
\newblock {\em Advances in Neural Information Processing Systems}, 2019.

\bibitem{rebuffi2017learning}
Sylvestre-Alvise Rebuffi, Hakan Bilen, and Andrea Vedaldi.
\newblock Learning multiple visual domains with residual adapters.
\newblock {\em arXiv preprint arXiv:1705.08045}, 2017.

\bibitem{Rebuffi2018EfficientPO}
Sylvestre-Alvise Rebuffi, Hakan Bilen, and Andrea Vedaldi.
\newblock Efficient parametrization of multi-domain deep neural networks.
\newblock {\em 2018 IEEE/CVF Conference on Computer Vision and Pattern
  Recognition}, pages 8119--8127, 2018.

\bibitem{rebuffi2017icarl}
Sylvestre-Alvise Rebuffi, Alexander Kolesnikov, Georg Sperl, and Christoph~H
  Lampert.
\newblock icarl: Incremental classifier and representation learning.
\newblock In {\em Proceedings of the IEEE conference on Computer Vision and
  Pattern Recognition}, pages 2001--2010, 2017.

\bibitem{rusu2016progressive}
Andrei~A Rusu, Neil~C Rabinowitz, Guillaume Desjardins, Hubert Soyer, James
  Kirkpatrick, Koray Kavukcuoglu, Razvan Pascanu, and Raia Hadsell.
\newblock Progressive neural networks.
\newblock {\em arXiv preprint arXiv:1606.04671}, 2016.

\bibitem{Rusu2016ProgressiveNN}
Andrei~A. Rusu, Neil~C. Rabinowitz, Guillaume Desjardins, Hubert Soyer, James
  Kirkpatrick, Koray Kavukcuoglu, Razvan Pascanu, and Raia Hadsell.
\newblock Progressive neural networks.
\newblock {\em ArXiv}, abs/1606.04671, 2016.

\bibitem{simon2021learning}
Christian Simon, Piotr Koniusz, and Mehrtash Harandi.
\newblock On learning the geodesic path for incremental learning.
\newblock In {\em Proceedings of the IEEE/CVF Conference on Computer Vision and
  Pattern Recognition}, pages 1591--1600, 2021.

\bibitem{Simonyan2015VeryDC}
K. Simonyan and Andrew Zisserman.
\newblock Very deep convolutional networks for large-scale image recognition.
\newblock {\em CoRR}, abs/1409.1556, 2015.

\bibitem{Ven2019ThreeSF}
Gido~M. van~de Ven and A. Tolias.
\newblock Three scenarios for continual learning.
\newblock {\em ArXiv}, abs/1904.07734, 2019.

\bibitem{wu2019large}
Yue Wu, Yinpeng Chen, Lijuan Wang, Yuancheng Ye, Zicheng Liu, Yandong Guo, and
  Yun Fu.
\newblock Large scale incremental learning.
\newblock In {\em Proceedings of the IEEE Conference on Computer Vision and
  Pattern Recognition}, pages 374--382, 2019.

\bibitem{yan2021dynamically}
Shipeng Yan, Jiangwei Xie, and Xuming He.
\newblock Der: Dynamically expandable representation for class incremental
  learning.
\newblock In {\em Proceedings of the IEEE/CVF Conference on Computer Vision and
  Pattern Recognition}, pages 3014--3023, 2021.

\bibitem{yin2020dreaming}
Hongxu Yin, Pavlo Molchanov, Jose~M Alvarez, Zhizhong Li, Arun Mallya, Derek
  Hoiem, Niraj~K Jha, and Jan Kautz.
\newblock Dreaming to distill: Data-free knowledge transfer via deepinversion.
\newblock In {\em Proceedings of the IEEE/CVF Conference on Computer Vision and
  Pattern Recognition}, pages 8715--8724, 2020.

\bibitem{2020More}
L. Yu, S. Parisot, G. Slabaugh, J. Xu, and T. Tuytelaars.
\newblock More classifiers, less forgetting: A generic multi-classifier
  paradigm for incremental learning.
\newblock {\em European Conference on Computer Vision}, 2020.

\bibitem{yu2020semantic}
Lu Yu, Bartlomiej Twardowski, Xialei Liu, Luis Herranz, Kai Wang, Yongmei
  Cheng, Shangling Jui, and Joost van~de Weijer.
\newblock Semantic drift compensation for class-incremental learning.
\newblock In {\em Proceedings of the IEEE/CVF Conference on Computer Vision and
  Pattern Recognition}, pages 6982--6991, 2020.

\bibitem{zenke2017continual}
Friedemann Zenke, Ben Poole, and Surya Ganguli.
\newblock Continual learning through synaptic intelligence.
\newblock In {\em International Conference on Machine Learning}, pages
  3987--3995. PMLR, 2017.

\bibitem{zhu2021prototype}
Fei Zhu, Xu-Yao Zhang, Chuang Wang, Fei Yin, and Cheng-Lin Liu.
\newblock Prototype augmentation and self-supervision for incremental learning.
\newblock In {\em Proceedings of the IEEE/CVF Conference on Computer Vision and
  Pattern Recognition}, pages 5871--5880, 2021.

\bibitem{zhu2021self}
Kai Zhu, Yang Cao, Wei Zhai, Jie Cheng, and Zheng-Jun Zha.
\newblock Self-promoted prototype refinement for few-shot class-incremental
  learning.
\newblock In {\em Proceedings of the IEEE/CVF Conference on Computer Vision and
  Pattern Recognition}, pages 6801--6810, 2021.

\end{thebibliography}
}

\end{document}